%% file: 0-main-sigconf.tex
\documentclass[sigconf,natbib]{acmart}  % required
\usepackage{xcolor} %for colored text
\usepackage[nomessages]{fp} %added by alex for colored bar in table
\usepackage{booktabs} % For formal tables
\usepackage{multirow} % multi-row in tables
\usepackage{url}
\usepackage{framed}
\usepackage{enumitem}  % itemize left space
\usepackage{balance}
\usepackage{siunitx} % align table at decimal points and using \num{}
\usepackage{amsmath} % greeks, math expressions
\usepackage{amsthm}
\usepackage{bm}  % for vector variables
\usepackage{algorithm}
\usepackage{algorithmic}
\usepackage{subcaption}
\usepackage{setspace} % (tom) from WSDM19, ACM suggestion for ACM reference format spacing
\usepackage{caption} % remove vertical spacing for sub-figure and sub-caption
\usepackage{soul}  % highlight text with \hl{}, and to strikethrough text by \st{}

\usepackage{graphicx}
\graphicspath{ {./images/} }

\usepackage{amsfonts}

\newlist{questions}{enumerate}{2}
\setlist[questions,1]{label=\textbf{RQ\arabic*},ref=\textbf{RQ\arabic*}}
\setlist[questions,2]{label=(\alph*),ref=\thequestionsi(\alph*)}

\definecolor{ForestGreen}{rgb}{0.13, 0.55, 0.13}
\definecolor{tomBlue}{HTML}{0196CE}
\definecolor{tomBlack}{HTML}{000000}

\definecolor{AkiRed}{rgb}{1.0, 0.3, 0.3}

\usepackage{xspace}

\AtBeginDocument{%
  \providecommand\BibTeX{{%
    \normalfont B\kern-0.5em{\scshape i\kern-0.25em b}\kern-0.8em\TeX}}}

\copyrightyear{2023}
\acmYear{2023}
\setcopyright{acmlicensed}\acmConference[SIGIR '23]{Proceedings of the 46th International ACM SIGIR Conference on Research and Development in Information Retrieval}{July 23--27, 2023}{Taipei, Taiwan}
\acmBooktitle{Proceedings of the 46th International ACM SIGIR Conference on Research and Development in Information Retrieval (SIGIR '23), July 23--27, 2023, Taipei, Taiwan}
\acmPrice{15.00}
\acmDOI{10.1145/3539618.3592031}
\acmISBN{978-1-4503-9408-6/23/07}

\begin{document}
\fancyhead{}

% \title{What Makes a Video Difficult for the Crowd to Classify?}
% \title{Are Citizen Scientists and Crowd Workers Complementary?}
\title{On the Impact of Data Quality on Image Classification Fairness}
% \affiliation{%

% \affiliation{%

\author{Aki Barry}
\email{aki.barry@uq.net.au}
\affiliation{%
%   \vspace{0.2em}
  \institution{The University of Queensland}
  % \city{City}
  \country{Australia}
}

\author{Lei Han}
\email{tomhanlei@hotmail.com}
\affiliation{%
%   \vspace{0.2em}
  \institution{The University of Queensland}
  % \city{City}
  \country{Australia}
}

\author{Gianluca Demartini}
\email{demartini@acm.org}
\affiliation{%
%   \vspace{0.2em}
  \institution{The University of Queensland}
  % \city{City}
  \country{Australia}
}

\renewcommand{\shortauthors}{Anonymized et al.}

\begin{abstract}
With the proliferation of algorithmic decision-making, increased scrutiny has been placed on these systems. This paper explores the relationship between the quality of the training data and the overall fairness of the models trained with such data in the context of supervised classification. We measure key fairness metrics across a range of algorithms over multiple image classification datasets that have a varying level of noise in both the labels and the training data itself. We describe noise in the labels as inaccuracies in the labelling of the data in the training set and noise in the data as distortions in the data, also in the training set. By adding noise to the original datasets, we can explore the relationship between the quality of the training data and the fairness of the output of the models trained on that data. 
\end{abstract}

\maketitle

\input{0-paper.tex}

\bibliographystyle{ACM-Reference-Format}
\balance
\bibliography{mybibliography}

% \clearpage

\end{document}

%% file: 0-paper.tex
\section{Introduction}
Fairness in the context of machine learning (ML) is an area that has seen increased interest in recent years and is still in its infancy \cite{Cat2020}. The understanding of fairness is extremely important in ML, as more important decisions are being made by algorithms rather than humans. Algorithmic decision-making is, for example, utilised in credit scoring \cite{Bac2017}, search  \cite{Swe2013}, and recidivism prediction \cite{Ang16} which are all subject areas in which fairness is an important factor \cite{Bar16}. In this work, we comprehensively measure the fairness of algorithms for image classification over a range of datasets and classification algorithms.

% Measuring fairness across this many dimensions will create a lot of data. Therefore, it is important to decide on which questions we would like answered. 
Our research can be summarised by the following questions:
\begin{questions}
  \item What is the relationship between noise in the labels and noise in the data in terms of the fairness of a classification model trained on such data? \label{itm:relationship}
  \item Do certain models achieve better fairness compared to others under noisy data conditions? \label{itm:models}
  \item Does transfer learning achieve better fairness across generalist/specialist datasets? \label{itm:transfer}
\end{questions}
Answering these questions will help guide decision-making on both the data and model selection when factoring fairness into account. The contributions that this paper make are: (i)~provide experimental results over different metrics of fairness across different models and datasets; (ii)~answer questions related to the impact of data quality on fairness (e.g., Does label accuracy increase fairness?); and (iii)~provide a starting point and datasets for future research into the impact of data quality on supervised classification fairness.

% \ab{Are these contributions okay?}

The key findings of this work include:
\begin{itemize}[topsep=0pt, noitemsep, leftmargin=*]
  \item Decreases in data quality almost always lead to lower demographic parity, accuracy parity, false-positive rate parity and calibration values while leading to higher false-negative rate parity values.
  \item Naive Bayes is largely immune to data imbalance as well as noise in the labels, when it comes to fairness.
  \item The complexity of the dataset seems to affect the importance of different data quality dimensions in terms of fairness.
\end{itemize}

% \ab{I feel like these findings could be hard to understand as the definitions aren't set yet. Should I make it vaguer?}

% This paper is organized as follows. Section 2 discusses similar research and takes a look at related literature. Section 3 discusses the methodology of the research. Section 4 shows some of the results over the used dataset as well as answers to the aforementioned research questions. Finally, Section 5 will discuss key insights gained from our results as well future possible research.

\section{Methodology}

\subsection{Measures of Fairness}
% The first question to address is how one is to define fairness.
% Although people would have a strong qualitative feeling as to what is fair and unfair,
% putting this to a quantitative number can be somewhat of a challenge.
% 
% On that note,
An important concept to consider when quantifying fairness is that of protected and unprotected attributes. Protected attributes are those which should ideally not have a bearing on the final decision \cite{Cat2020}. These can include but are not limited to sex, race, gender, and religion. For consistency's sake, we introduce the following notation in our work:
% $$\begin{array}{ll}
(i)~$X_p$: protected attributes; (ii)~$X_u$: unprotected Attributes; (iii)~$Y$: correct decision; (iv)~$\hat{Y}$: model's prediction of correct decision.
With this notation, in the following, we introduce commonly used definitions of fairness, namely anti-classification, classification parity, and calibration.

\textit{Anti-Classification} is defined as not using the protected attributes in the model \cite{Cor18}. This means that the results of the model should remain the same for different protected variables if the unprotected variables are the same. This can be defined as: $\hat{Y}(X_p, X_u) = \hat{Y}(X_u)$.
% Although this by itself may seem like the ideal solution to fairness, 
With this definition, some unprotected attributes may be correlated with protected attributes, such as postcode and race, and so this is typically not a bulletproof solution. Also, it has been found that models enforcing this rule can actually have negative effects on the classes that they aim to protect \cite{Cor18}. It should also be noted that in the context of image recognition, the protected variable is often what is predicted.

\textit{Classification Parity} is a condition that aims to have parity across protected classes for some kind of measure. Common examples include demographic parity, accuracy parity, false-positive rate (FPR) parity, and false-negative rate (FNR) parity.
Demographic parity is when the positive decisions remain the same across protected attributes. This can be defined as: $\mathbb{P}(\hat{Y} = 1 | X_p) = \mathbb{P}(\hat{Y} = 1)$.
% There are some obvious limitations to enforcing this such as the fact that there could exist different underlying positive rates between protected demographics which would be equalized away under this definition. An example could be that of classifying the type of insect eaten by an anteater. It would be odd if ants had the same rate of prediction as flies \cite{Gal2017}.
% 
Accuracy Parity is when the accuracy across classes is the same. This can be defined as: $\mathbb{P}(\hat{Y} = Y | X_p) = \mathbb{P}(\hat{Y} = Y)$.
This metric has upsides compared to demographic parity as it no longer assumes that the underlying prevalence\footnote{Prevalence is the proportion of the dataset that is a specific class.} is the same. However, it does not distinguish between false positives (FP) and false negatives (FN) which could be unwanted behaviour.
% (Perhaps FPs are more costly than FNs).
% 
FPR parity implies that the FP rate should be equal across protected attributes, or formally
% Put formally, this can be described as:
% 
$\mathbb{P}(\hat{Y} = 1 | Y = 0, X_p) = \mathbb{P}(\hat{Y} = 1 | Y = 0)$.
Similarly for FNR parity we have:
% \[\mathbb{P}(\hat{Y} = 0 | Y = 1, X_p) = \mathbb{P}(\hat{Y} = 0 | Y = 1)\]
$\mathbb{P}(\hat{Y} = 0 | Y = 1, X_p) = \mathbb{P}(\hat{Y} = 0 | Y = 1)$.
%
% For a classification task, a FP would indicate the identification of a specific class that is incorrect while a FN would be an incorrect identification of a class when it is another class.
In the context of classification, this would mean that the proportion of incorrect positive/negative identifications should be equal across all classes. However, focusing on just this metric can lead to accuracy degradation and if the underlying rates are different then some level of imbalance in the error rates must exist \cite{Cho2016}.
It should be noted that there exists a relationship between FPR and FNR and this is encapsulated by:
% the following equation:
% 
$FPR = \frac{p}{1-p}\frac{1-PPV}{PPV}(1-FNR)$,
where \(p\) is prevalence, \(PPV = \frac{TP}{TP + FP}\), \(TP = \text{True Positives}\), and \(FP = \text{False Positives}\). If both FPR and FNR are equal across groups, then a model that is independent across protected attributes cannot exist unless the prevalences are the same \cite{Cho2016}. 

\textit{Calibration} is an adjustment that is made such that the expected prediction per group is approximately equal to the underlying prevalence in the training data \cite{Heb2017}. Formally, this can be described as:
$\mathbb{P}(Y=1 | \hat{Y}, X_p) = \mathbb{P}(Y = 1 | \hat{Y})$.
This means that the proportion of actual positives given a positive or negative prediction should be the same across protected attributes.
Then,
% with all these definitions we should just adjust every algorithm to apply all of these different fairness metrics to get a ``super fair'' result. This is likely impossible as
the only way that calibration, FPR, and NPR can all be upheld at once is if the predictor is perfect, or, the base rates of the protected attributes are the same  \cite{Kle2017}. Both of these scenarios are unlikely in real-world data. So effectively only two of these conditions can be upheld at once. This holds similarly for other combinations of fairness metrics.
% and so it is impossible to satisfy all fairness criteria at once.

% A final topic worth exploring is what inherently causes unfairness in machine learning. To this end
Further, there exists a consensus on what can cause unfairness in the models, including biases encoded in the data, the fact that maximising accuracy fits the majority, and the need to explore outcomes \cite{Cho2018}. 
% 
% The most obvious is the bias encoded in the data.
% In computing, garbage in, garbage out is a common idiom. That is, 
% Usually, if the input for a particular algorithm is suboptimal then the output will also be suboptimal. 
% Similarly, biases in the data will be encoded into the machine learning model.
% 
When maximising accuracy in any dataset, if the unprotected variables between two groups with different protected attributes have a different underlying distribution concerning the decision, a classifier following anti-classification will inevitably fit the majority group to maximise accuracy. This has the effect of advantaging some groups and disadvantaging others. 
% 
% For data to be created it must first occur in the real world. In this way, 
For critical decisions, in particular, there is less data available for seemingly suboptimal decisions. Thus, a fully unbiased view of the decision space cannot be had without considerable cost. We emulate this in our experiment by making the data unbalanced for specific classes and then measuring the effects on the models' fairness.

\subsection{Definitions of Data Quality}
% When it comes to defining data quality things become more established.
Data quality issues related to fairness can be split into two main sub-groups, those being data coverage and data noisiness. Coverage can be qualitatively described as the extent to which the observed population is representative of the overall population \cite{Koh1998}.
% Although there are a few ways this can be defined, we will use equivalence partitioning.
\textit{Equivalence Partitioning}, a way to define this, can be described as the prevalence of a specific class divided by the total number of classes \cite{Man2019}. This can be defined as:
$EP_i = p_i \cdot N_c$,
where \(EP_i\) is the equivalence value of a class \(i\), \(p_i\) is the prevalence of a class \(i\), and \(N_c\) is the total number of classes in the training set. In a balanced dataset the equivalence partitioning is one for all classes. This value can thus be used to quantitatively measure the `balanced-ness' of a specific class in the data.
There are two types of noise: that of the labels and that of the data itself. For noisy labels, a simple metric we can use is the class-wise accuracies of the labels. To introduce noise in the data, we use two popular techniques: Uniform noise over the whole image, that is, a random change in brightness for some proportion of the data, and gaussian noise, which is the pixel-wise gaussian noise common in images shot under low-light conditions.

\subsection{Datasets}
% With the previous definitions complete, we must look at which datasets to use for our analysis. To this end, datasets with high levels of fairness needed to be chosen to have a good baseline. To this end, statistical tests have been performed on popular machine learning image classification datasets \cite{Man2019}. A research paper has performed a covariate shift test as well as an equivalence partitioning test on a range of benchmark vision datasets. A covariate shift test is that given a model \(\mathbb{P}(Y|X)\mathbb{P}(X)\) how does \(\mathbb{P}(X)\) differ between training and test sets
\citet{nair2019covariate} compared the MNIST, FMNIST, CIFAR-10, CIFAR-100, and SVHN datasets and found that the covariance shift in the MNIST dataset was the greatest.
% and so this dataset was initially ruled out.
For the equivalence partitioning test, all the datasets had balanced datasets except for SVHN.
% Therefore, the remaining pool of eligible datasets would be the FMNIST, CIFAR-10, and CIFAR-100 datasets.
For these reasons, the remaining three datasets will be used in our investigation. FMNIST is a domain-specific dataset while CIFAR-10 and CIFAR-100 (with coarse labels) are general domain datasets. We additionally consider a second domain-specific dataset that differentiate individual grains of rice into 5 different species \cite{Cin2019}.
% It should be noted that the CIFAR-100 dataset was used with the coarse labels rather than the fine labels which are less specific. An example of this is replacing the `beaver', `dolphin', `otter', `seal', and `whale' labels with just the `aquatic mammals'.

% \subsection{Algorithms Compared}
% The final piece of the puzzle will be the algorithms that will be used to test. In this regard, a large selection of algorithms were chosen which aimed to cover the largest amount of methods. These were, in no specific order, k-nearest neighbours (k-NN), logistic regression, support vector machine (SVM), Naive Bayes (NB), decision tree, gradient boosted tree, regular neural network, convolutional neural network (CNN), and pretrained neural network. For the k-NN, logistic regression, SVM, NB, decision tree, and gradient boosted tree, a feature vector was calculated from the images and fed to the algorithms. The feature vector was calculated by using the EfficientNetV2B0 neural network with the weights being pre-trained on ImageNet and the output layer removed. In terms of the pretrained models being tested in this report, the VGG-16 model as well as the ResNet50V2 model was used. The weights on these were also pretrained on ImageNet.

\begin{figure}[t]
  \centering
  \includegraphics[scale=0.45]{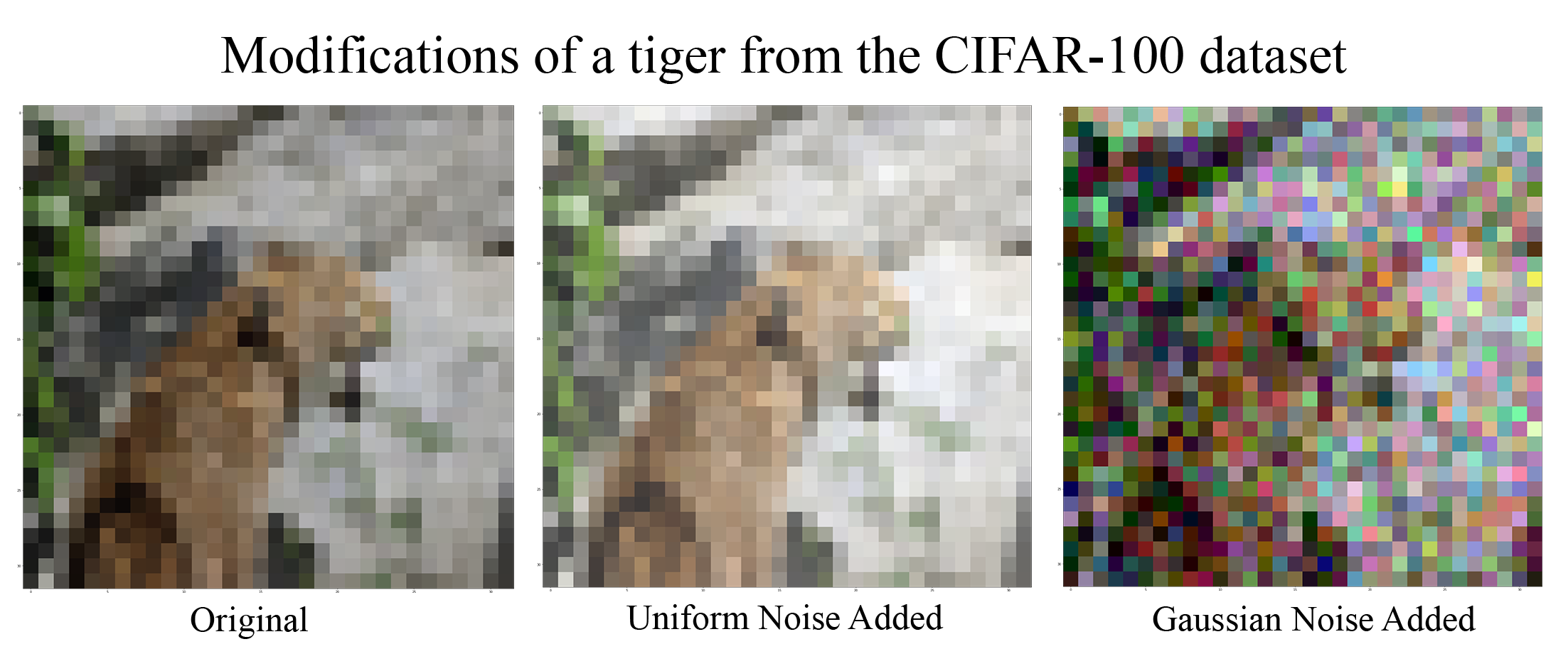}
  \caption{Examples of uniform and Gaussian noise.}
  \label{fig:example_tiger}
\end{figure}

% \subsection{Dataset Modification}
We modify the original datasets to introduce noise in a controlled fashion.
% The possible parameters to control for were equivalence partitioning, label accuracy, the proportion of uniform noise, and the proportion of Gaussian noise. These parameters would be modified for a singular class to create new datasets to train on.
%
For \textit{equivalence partitioning}, each dataset was modified such that the equivalence partitioning value of the target class would equal a specific value while the rest of the classes each have an equal amount to pick up the slack.
% For example, for a dataset with 10 classes with 10 data points each and a target EP value of 0.1, the target class would have all but one of the data points removed. This can be seen by the below calculation.
%
% \[\text{EP value of \(x\)} \iff x \approx \frac{N_c \cdot C_d}{N_d}\]
%
% where \(N_c\) is the number of classes, \(C_d\) is the number of points in the targeted class, and \(N_d\) is the total number of points in the dataset. Therefore, in the aforementioned case, the EP value would technically be \(0.10989 \approx \frac{10*1}{91}\). However, this is as close as we can get without having to throw out data from other classes.
%
For \textit{label accuracy}, a proportion of the target class labels are set to a uniform randomly chosen other class. For example, for a label accuracy of 50\%, half of the targeted class labels would be changed to any of the other classes with an equal probability.
For \textit{uniform noise}, the value given will be the proportion of the specific class without added uniform noise. So a value of 0.6 would indicate that 60\% of the data in the class will remain unmodified while 40\% will be. The added noise is drawn from a uniform distribution from 50 to 60 and this was added to/subtracted from the brightness of the image
(with a max of 255 brightness). 
% (max$=$255). 
% These values were chosen by eye as they were perceived to be in the range of brightness differences seen in real-world data.
% In future, more scientific rigour could be applied to value selection however that was out of scope for this report.
%
For \textit{Gaussian noise}, the brightness of each pixel is set from a normal distribution
% with a mean of 0 and a standard deviation of 30. 
($\mu$$=$$0$, $\sigma$$=$30).
% These values were chosen by eye as they looked similar to noisy images seen in real datasets.
% Again, more scientific rigour could be applied to the value selection but once again that is not really within the scope of this paper.
The Gaussian noise value will be the proportion of the data in the class that remains unmodified so a value of 0.2 will have more noisy data points than a value of 0.4. Figure~\ref{fig:example_tiger} shows examples of both types of noise.

% What is interesting with this comparison is that even though the expected standard deviation of the values of each pixel in the uniform noise is almost double that of the gaussian noise, our eyes interpret the Gaussian noise as much more undecipherable and this was also reflected in the results of our experiment.

\subsection{Experimental Setup}
% With all the parameters and possible variables laid out, the actual experimental methodology can be explained. 
We compare the following classification algorithms: k-nearest neighbours (k-NN), logistic regression, support vector machine (SVM), Naive Bayes (NB), decision tree, gradient boosted tree, regular neural network, convolutional neural network (CNN), and pretrained neural network. For the first six, a feature vector was calculated from the images and fed to the models. The feature vector was calculated by using the EfficientNetV2B0 neural network with weights being pre-trained on ImageNet and the output layer removed. In terms of pretrained models, we use VGG-16 as well as ResNet50V2. Their weights were also pretrained on ImageNet.

We train all these algorithms on the aforementioned datasets and we measure fairness  by demographic parity, accuracy, FPR, FNR and calibration. The class-specific values of these fairness metrics were calculated and then reduced to a single number. For demographic parity and calibration, this number was simply calculated as the class-specific value subtracted by one. In this way, a value below zero would mean the class was underrepresented in the predictions while a value above zero would mean the class was overrepresented. For the other metrics, the fairness value was calculated to be the value of the metric for that specific class subtracted by the mean of that metric for all the other classes. In this way, a value closer to zero would mean increased fairness. This is written as: $\text{ParityValue} = \text{ClassValue}[t] - \frac{\sum_{i \neq t} \text{ClassValue}[i]}{N_c - 1}$,
where \(t\) is the target class and $N_c$ is the total number of classes. Therefore, if the target class had an accuracy of 50\% and the other classes all had an accuracy of 70\% the parity value would be $-20\%$. These values were calculated for data quality of 0.2, 0.4, 0.6, 0.8, and 1.0 for all the data quality variables individually, and for 0.33, 0.66, and 1.0 for all combinations of our data quality measures. 
% GD bring back if space:
% For example, these models were run on a version of the dataset with a label accuracy value of 0.4, and these models were also run on a version of the dataset with a label accuracy value of 0.33 as well as an equivalence partitioning value of 0.33. 
Each experiment was repeated three times and results averaged.
% each on different classes for each of the modified instances to try and iron out any inherent biases in the data.

\section{Results and Discussion}
% After running these models across all the possible combinations of datasets and modifications to the datasets, we focused on answering the aforementioned research questions by analysing the large troves of collected data.

\subsection{Fairness with Noisy Labels (\ref{itm:relationship})}

\begin{figure}[t]
  \centering
  \includegraphics[width=0.47\textwidth]{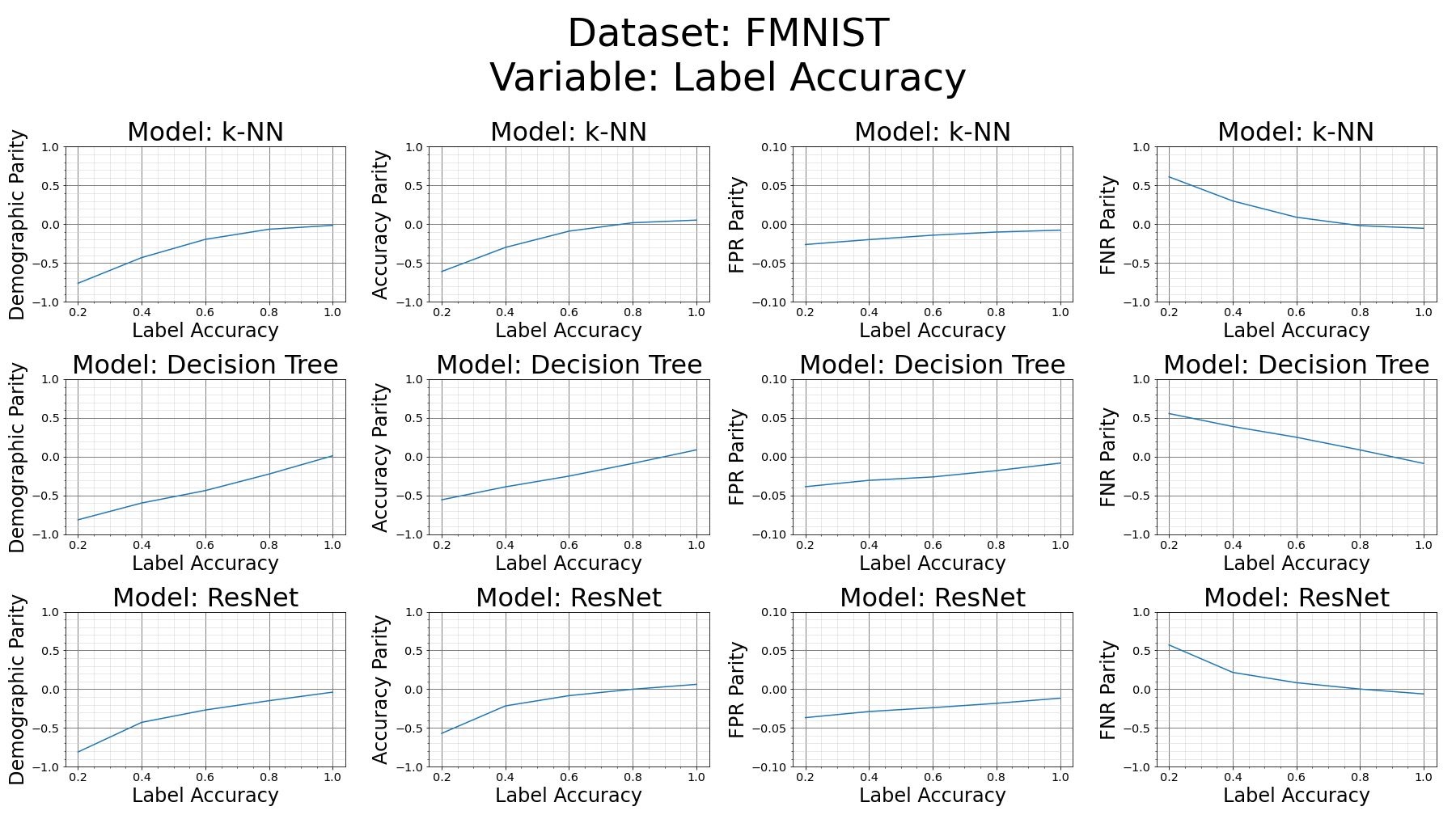}
  \caption{Improvement of fairness over training data quality (i.e., varying levels label noise) on the FMNIST dataset (Note: Some models and measures were omitted due to space constraints).}
  \vspace{-2mm}
  \label{fig:noise}
\end{figure}

Figure~\ref{fig:noise} shows the level of fairness over training data quality levels across all considered metrics.
In the FMNIST dataset for noisy labels, we see that the values for demographic parity, accuracy parity, FPR parity, and calibration were mostly monotonically increasing from less quality to more quality while the FNR parity was mostly monotonically decreasing.
% The graph for this is below.

We observed the same pattern across all of data quality metrics and datasets.
This is opposite to our intuition that the quality for a certain class gets worse the false positive rate increases.
% This pattern is interesting as we would expect that as the quality for a certain class gets worse the false positive rate would increase, however, we saw the opposite.
This can be explained
% with an interesting theory though, across almost all of the models we tested,
as the quality of a specific class decreases, the model predicts that class a lot less. This
% describes the results we see and would
implies that `fitting to the majority' seems to be the strongest behaviour we see from these models.
% Also, note that the datasets in which the fairness is not zero for the baseline are more a product of the bias inherent in the training dataset rather than the models themselves.
Also, Naive Bayes seems to be an outlier in that it is the only model that is resilient to certain types of noise in the data.
% This will be explored further in later section.

\begin{figure}[t]
  \centering
  \includegraphics[width=0.47\textwidth]{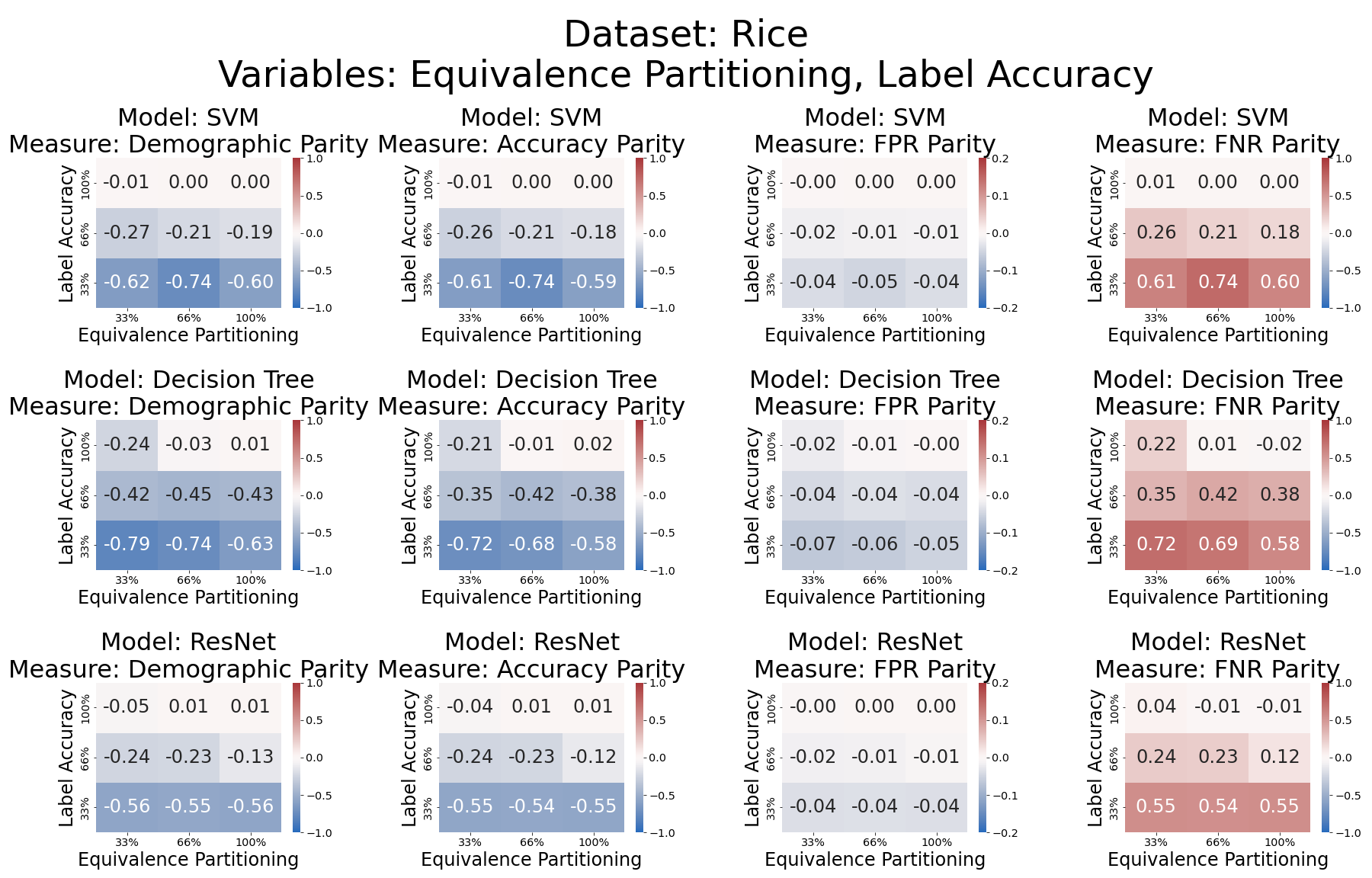}
  \caption{Fairness metrics over training with noisy labels, on the rice dataset.}
  \vspace{-1mm}
  \label{fig:rice}
\end{figure}

We also see that for the domain-specific FMNIST and rice datasets, a low label accuracy is more detrimental to the fairness of the algorithm when measuring equivalence partitioning. For example, a label accuracy of 33\% had an accuracy parity value $-59\%$ for SVM meaning that the class that had inaccurate labels was predicted incorrectly 59\% worse than every other class. When compared to the same dataset with an equivalence partitioning value of 33\%, SVM had an accuracy parity value of $-1\%$. This was observed across all models and all metrics for these datasets and shows that the poor label accuracy was much more detrimental to the results. The results for the rice dataset are shown in Figure~\ref{fig:rice}. %\gd{here or in the caption goes a description of what the fig shows. also, it's too small to read, so maybe drop some matrices and show only a few of them?}
This effect was less pronounced for the CIFAR-10 and CIFAR-100 datasets where we saw a more symmetrical relation between label accuracy and equivalence partitioning. A possible interpretation is that the models are challenged by the more complex general-domain classification task and so training data quality is much more important for fairness. 
This is also supported by the non-linear nature in which fairness improves across different equivalence partitioning values in the FMNIST dataset, whereas we see a much more linear trend in the CIFAR-10 and CIFAR-100 datasets. Also, the micro-average \(F_1\) scores in the  CIFAR-10 and CIFAR-100 datasets for the k-NN model are 49.46\% and 38.82\% respectively while on the FMNIST and rice datasets the same model had \(F_1\)  of 83.42\% and 99.2\% respectively. Clearly, a change in equivalence partitioning would affect the models built on these datasets a lot less. 

\subsection{Fairness with Noisy Data (\ref{itm:models})}

When it comes to noise in data, we saw that the Gaussian noise contributed  more to unfairness than  uniform noise, even though the values of the brightness of individual pixels in the uniform noise had a greater standard deviation from the original value. This could be attributed to the image-to-feature vector translation which may inherently be more flexible with different brightnesses. Other than that, the results were very similar to what we saw with the equivalence partitioning and label accuracy scenario with most of the unfairness being contributed to by the Gaussian noise rather than the uniform noise even on the more difficult CIFAR datasets.
The  fairness metrics we considered were all generally extremely correlated when controlling for dataset, model, and data quality metric, except for FNR parity which was negatively correlated.

We found that Naive Bayes (NB) is largely resistant to changes in class balance in the training set (equivalence partitioning), as well as label accuracy.
% This result was quite surprising to us and we found that
% In terms of the metrics of data quality, Naive Bayes was the only model that we tested that had this resistance to these kinds of unfairness.
%
Our explanation is that due to the way in which NB calculates different distributions for each of the classes, the effect of one poor class does not affect the overall model. 
In the equivalence partitioning case, we can assume that the predicted distribution for the modified class will likely not be that different, and so the outcome for the model would not be significantly affected. 
In the accuracy modification case, we would see that each of the other classes would have their distributions equally move toward the poor class and so these effects would largely cancel each other out. 
This interpretation is supported by measuring the overall \(F_1\) score across different versions of a dataset. For equivalence partitioning values of 0.2, 0.4, 0.6, 0.8, and 1.0, on the CIFAR-100 dataset, we observed \(F_1\) scores of 47.99\%, 47.92\%, 48.01\%, and 48.04\% respectively. For label accuracy values of 0.2, 0.4, 0.6, 0.8, and 1.0, on the CIFAR-100 dataset, we observed \(F_1\) scores of 47.68\%, 47.9\%, 47.9\%, 47.98\%, and 48.04\%.
The model  achieves similar results even with only 20\% label accuracy for one of the classes and this is reflected in the almost unchanged values  of  fairness.
The downside of using NB to maximise fairness is that with Gaussian noise NB often displays the worst fairness among all the models we considered.

In terms of algorithms that are most resistant to noise in the data, there was not a clear winner compared to label accuracy and equivalence partitioning. We saw that k-NN, neural network, CNN, as well as both of the transfer learning neural networks seemed to perform well. However, this was not consistent over all of the datasets and the ordering of these algorithms in terms of fairness.
%
% Therefore, we believe that while neural networks and k-NN might be somewhat resistant to noise in the data, larger amounts of research will need to be done to get a more statistically significant result.
%
In summary, 
% with respect to which algorithm would be the best for some combination of noise in the data as well as noise in the labels, 
we found that the best algorithm to deal with noise in data and labels varied across datasets and recommend that, if only label accuracy and equivalence partitioning issues exist, Naive Bayes should be used when there is a need to prioritise fairness.
% if fairness is the only thing important, otherwise a comparison will have to be done.

\subsection{Transfer Learning and Data Quality (\ref{itm:transfer})}

% Due to the difference in difficulty between all of the datasets, apples-to-apples comparisons could not be made if transfer learning would have better fairness for specialist or generalist datasets. So,
We compare the performance of the transfer learning models and the other neural network models we tested.
% Although this is not a perfect comparison, fully training the same neural network on our limited hardware would be infeasible.
We found that relative to the regular neural network and convolutional neural network, transfer learning on specialist datasets (FMNIST and rice) showed an average improvement of $-6.5\%$ and $-3.5\%$ respectively in accuracy parity with versions of the dataset with 33\% label accuracy and 33\% EP values. On the generalist datasets (CIFAR-10 and CIFAR-100) transfer learning had an average improvement of 11.5\% and 4.5\% respectively.
This may due to the complexity of
% Although this looks good for the case that transfer learning seems to be an improvement, since
the CIFAR datasets.
% are the most complicated, this could be the variable affecting the results rather than the fact that the results are generalist.
The transfer learning models are however not significantly better than any other of the models that were tested, in terms of the fairness metrics.

\section{Related Work}
There is limited research on the direct effect of data quality on fairness. The most related work is that of looking at the effect of data quality on non-class specific metrics such as overall accuracy and $F_1$ score \cite{Bla2011,Bud2022}. 
Other recent work has looked at fairness in ML, as well as at the impact of data quality on ML, unrelated to fairness.

\textit{Fairness in Machine Learning.}
% Fairness in machine learning is an important topic that has seen increased interest throughout the years. 
% With algorithmic decision-making being at the helm for large amounts of decision-making in the field, scrutiny must be placed on these systems to ensure that they are fair. 
% 
There are a number of solutions to maintain fairness in ML models \cite{kamiran2010discrimination,vzliobaite2011handling,kearns2018preventing}. These are typically based on the principle of removing (part of) the effect of bias and discrimination in the training data so that the classifiers learnt from this dataset are fair.
Previous research has also attempted to describe fairness in algorithms that consist of two steps: (i)~defining a set of criteria, and (ii)~developing a decision rule to satisfy the criteria \cite{corbett2017algorithmic}. Others have defined their approach by using probability expressions, such as \cite{feldman2015certifying,hardt2016equality,pleiss2017fairness,zafar2017fairness}.
% 
% On the other hand, the `blind spot' problem in ML calls for the improvement of algorithms to keep the fairness of automatic approaches. 
According to the interviews conducted by Holstein et al.~\cite{holstein2019improving}, people's specific needs being sufficiently expressed in ML solutions has raised considerable concerns. This highlights the importance of maintaining a balanced dataset to develop any ML algorithm.

\textit{Impact of Data Quality on Machine Learning.}
Existing research has looked at mitigating bias in ML models. For example, Hube et al.~\cite{hube2020debiasing} proposed a debiasing approach when creating embeddings by
% Their method
incorporating the loss function in the training process
with a pre-trained binary (e.g., positive or negative) classifier.
% In this way, the bias of the trained embeddings can be mitigated.
% 
Boulitsakis-Logothetis~\cite{Bou2022} proposed a modification to the Naive Bayes algorithm that gives guarantees on the statistical parity across protected attributes.
% This algorithm involves first splitting the dataset into subsets with different protected attributes and then training these individually, it then combines these sub-models in a way that would enforce the desired statistical parities.
This paper found benefits in terms of increases in fairness measures across protected classes but found that accuracy would often be degraded.
Bechavod et al.~\cite{Bec2017} proposed a method of penalizing unfairness which was inspired by regularization. In this way, they have introduced a data-dependent penalty to the learning process. They found good results balancing between overall accuracy and fairness and performed better than other fairness methods \cite{zafar2017fairness}.
As compared to previous work, we looked at the impact of noisy training data on the level of fairness achieved by models trained with such data, compared the robustness to noise of different models, and also looked at a transfer learning scenario.

\section{Conclusions}
Our work has addressed questions about the relationship between fairness and data quality. We quantitatively measured fairness across a range of definitions of fairness and data quality. We have made key observations on the relationship between data quality and fairness.
% Also, we have seen the outcomes of different models in search of an algorithm that is resistant to changes in data quality in terms of fairness. Here 
We observed the advantage of Naive Bayes in terms of its ability to still give fair results with differing levels of label accuracy as well as equivalence partitioning.
We found that Gaussian noise has a much larger impact on fairness on a per standard deviation basis when compared to uniform noise.
Finally, we checked if transfer learning had some kind of edge when it came to fairness over both generalist and specialist datasets, and found that they do not. 

%\gd{add a sentence about gaussian vs uniform noise differences?}

% For future work, we aim to have a more realistic comparison by using a grid search for each model and then compare the best one as this would better match what the process would be like in the real world, look at different ways in which we could modify the data besides gaussian and uniform noise, and look at a problem besides image classification to hopefully see that our results carry over.

%\gd{make a pass over references and add publication venue (journal of the conference) if missing and available.}

% \noindent
\textit{Acknowledgments.}
This work is partially supported by the Australian Research Council (ARC) Discovery Project
under Grant No. DP190102141, the ARC Training Centre for Information Resilience (Grant No. IC200100022), and the Swiss National Science Foundation (SNSF) under contract number CRSII5\_205975.